\documentclass[runningheads]{llncs}
\usepackage[T1]{fontenc}
\usepackage{graphicx}
\usepackage{amsmath,amssymb,amsfonts}
\usepackage{xcolor}
\usepackage{float}
\usepackage{hyperref}
\usepackage{tabularx,booktabs}

\begin{document}
\title{SoccerKDNet: A Knowledge Distillation Framework for Action Recognition in Soccer Videos}

\author{}
\author{Sarosij Bose\inst{1}\orcidID{0000-0003-3014-4796} \and
Saikat Sarkar\inst{2}\orcidID{0000-0001-7118-6058} \and
Amlan Chakrabarti\inst{3}\orcidID{0000-0003-4380-3172}}
\authorrunning{Sarosij Bose et al.}
%
\institute{}
 \institute{Department of Computer Science and Engineering, University of Calcutta, India\\
 \and
 Department of Computer Science, Bangabasi College, University of Calcutta, India\\
 \and
 A.K.Choudhury School of Information Technology, University of Calcutta, India\\
 \email{\{sarosijbose2000, to.saikatsarkar17\}@gmail.com}, \email{acakcs@caluniv.ac.in}}

\titlerunning{SoccerKDNet}
\maketitle
\begin{abstract}
Classifying player actions from soccer videos is a challenging problem, which has become increasingly important in sports analytics over the years. Most state-of-the-art methods employ highly complex offline networks, which makes it difficult to deploy such models in resource constrained scenarios. Here, in this paper we propose a novel end-to-end knowledge distillation based transfer learning network pre-trained on the Kinetics400 dataset and then perform extensive analysis on the learned framework by introducing a unique loss parameterization. We also introduce a new dataset named "SoccerDB1" containing 448 videos and consisting of 4 diverse classes each of players playing soccer. Furthermore, we introduce an unique loss parameter that help us linearly weigh the extent to which the predictions of each network are utilized. Finally, we also perform a thorough performance study using various changed hyperparameters. We also benchmark the first classification results on the new SoccerDB1 dataset obtaining 67.20\% validation accuracy. The dataset has been made publicly available at: \textcolor{magenta}{\url{https://bit.ly/soccerdb1}}

\end{abstract}

\section{Introduction}
\label{sec:intro}
Recognition of player actions in soccer video is a challenging computer vision task. Existing vision-based soccer analytics models either rely heavily on manpower which is responsible for tracking every aspect of the game or other offline network based analytics products that are used for analyzing the game closely once it's over \cite{sarkar2022soccerpr}, \cite{sarkar2018estimation}. It has been established that deep learning based methods exceed their traditional counterparts in performance. Recently, deep reinforcement learning based models \cite{9840895}, \cite{sarkar2022watch} have been used for the estimation of ball possession statistics in broadcast soccer videos. 
But, there is an issue regarding employing such deep networks, which are often trained on large image based datasets such as ImageNet. These offline models may deliver superior accuracy but suffers due to a significant domain gap. As a result, there is a need for domain specific data or, at the very least, fine tuning on sports specific datasets.

\par \textbf{Soccer Action Recognition:} We also look into the existing literature in the action recognition domain for soccer videos. However, work has been very limited regarding this aspect. One of the very few public soccer video datasets is the Soccernet v2 benchmark \cite{deliege2021soccernet} released very recently. Other attempts to classify actions from soccer videos such as in \cite{cioppa2020context} have focused on specific localization tasks instead of classification. Therefore, we believe that our newly contributed soccer dataset (SoccerDB1) and knowledge distillation based action recognition framework (SoccerKDNet) would help progress the research on vision-based action recognition in soccer videos. 

In summary, we present SoccerDB1, a datset for action recognition in soccer video. We also present SoccerKDNet, a knowledge distillation based action recognition framework. We have achieved 67.20\% accuracy on action recognition task. Next, we describe our SoccerDB1 dataset in detail.

\section{SoccerDB1 Dataset}
\label{sec:db}
\par
\begin{figure*}[ht!]
		\centering
		\vspace{20pt}%
		\includegraphics[scale=0.45]{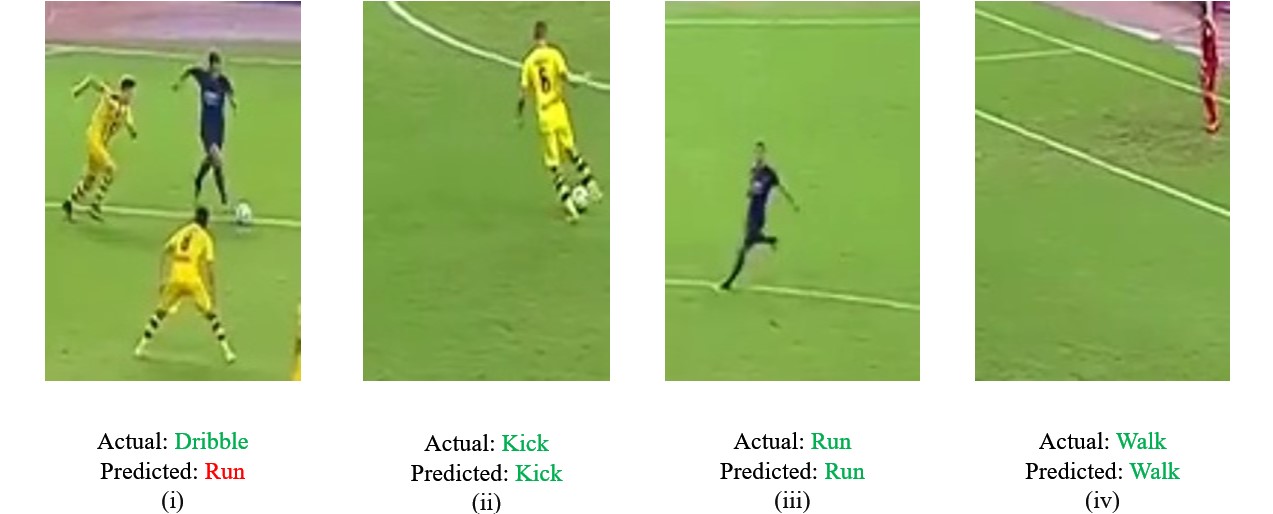}
		\caption{\label{fig:datasetcollage} Sample frames from SoccerDB1 dataset, their actual and predicted labels.}
\end{figure*}

 We introduce a new soccer dataset named SoccerDB1 consisting of 448 soccer video clips. The dataset contains videos of 4 action classes namely: Dribble, Kick, Run, and Walk. There are over 70 video clips per class. 
Sample frames of different action classes are shown in Figure \ref{fig:datasetcollage}.
The video clips are created manually from openly available broadcast soccer match videos available on YouTube. The frames were sampled uniformly and each video clip contains 25-26 frames. The proposed action recognition framework is discussed next.

\section{Methodology}
\label{sec:methodology}
\begin{figure*}[ht!]
		\centering
		\includegraphics[height=0.4\linewidth, width=\textwidth]{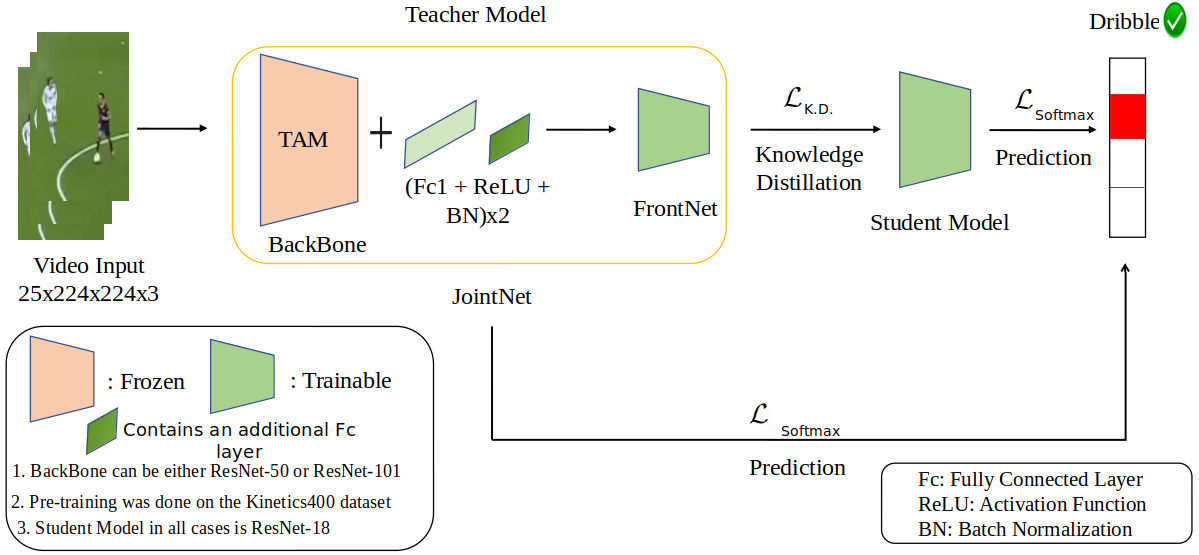}
		\caption{\label{fig:fullarch} SoccerKDNet: Schematic Architecture of Proposed End-to-End network.}
	\end{figure*}
 
\subsection{Knowledge Distillation based Transfer Learning}






\par
\begin{figure}[ht]
 		\centering
 		\frame{\includegraphics[height=0.4\linewidth, width=0.9\linewidth]{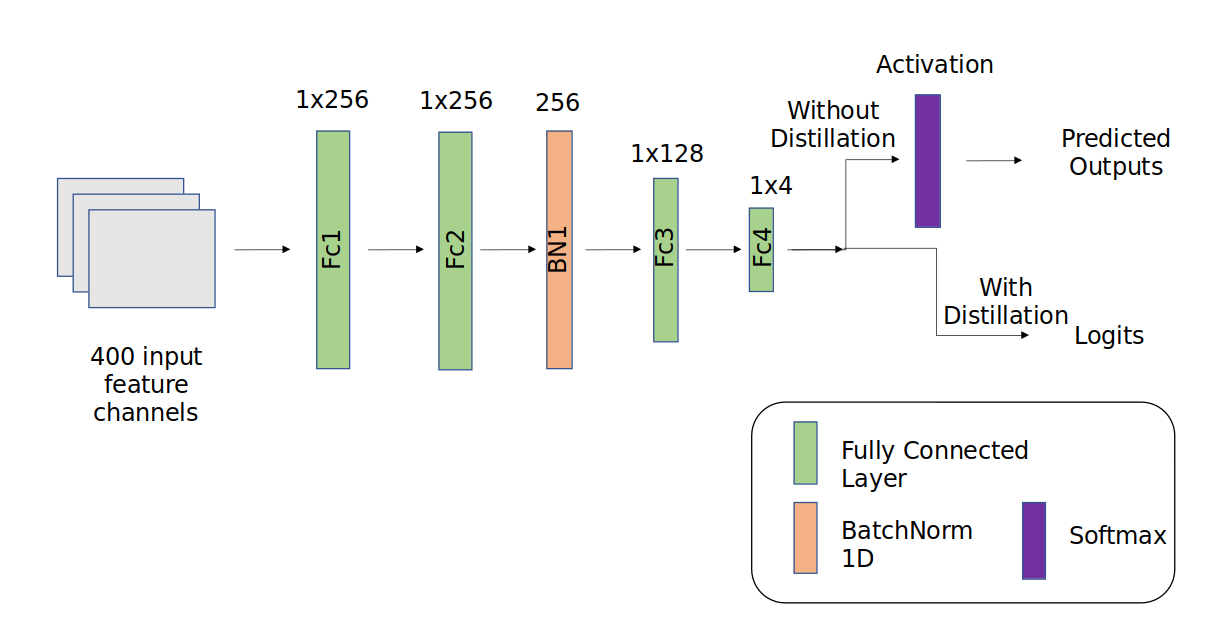}}
 		\caption{Architecture of the FrontNet Module. The Backbone network in Figure \ref{fig:fullarch} has an output of 400 classes so the obtained feature output serves as it's input in this case.}
\label{fig:frontnet}
 \end{figure}

 We propose the SoccerKDNet network to classify actions from soccer video clips as shown in Figure \ref{fig:fullarch}. Here, we use the Temporal Adaptive Module (TAM) \cite{liu2021tam} with backbones as both ResNet-50 and ResNet-101 \cite{he2016deep}. We then add a few fully connected layers with BatchNorm in front of the backbone network in order to enable it to have some learnable parameters. The features from this setup are then passed on to the frontnet which is shown in Figure \ref{fig:frontnet}. This entire setup is referred to as the 'jointnet' throughout the rest of the paper. 
\par We use ResNet-18 as the student network in all our experiments. The 'jointnet' serves as the Teacher Network and is initially trained on the Soccer Dataset. We use both ResNet-50 and ResNet-101 as backbones alongwith the Temporal Adaptive Module (TAM). We perform uniform sampling with all the video frames in all our experiments since it is known to yield better results than dense sampling \cite{chen2021deep}.

\par In our network as shown in Figure \ref{fig:fullarch}, the various losses employed are described below:-

\begin{itemize}

\item Cross Entropy Loss:-\\
The Cross Entropy Loss is given by the following formula:-
\begin{equation}
\label{eq:formula1}
\mathcal{L}\textsubscript{softmax}= -\sum_{c=1}^My_{o,c}\log(p_{o,c})
\end{equation}
\item KullBack-Liebler Divergence Loss:-\\
The KullBack Liebler Divergence Loss can be given by the following formula:-
\begin{equation}
\label{eq:formula3}
 \mathcal{L}\textsubscript{KL} = \sum_{c=1}^{M}\hat{y}_c \log{\frac{\hat{y}_c}{y_c}}
\end{equation}
We have also applied a Temperature ($\tau$) hyperparameter to this equation.
\item Knowledge Distillation Loss:-\\
If there is a given Teacher Network D and a Student Network S and the loss of the student network is denoted by $\mathcal{L}\textsubscript{softmax}$ and a hyperparameter $\alpha$ such that ($0\leq\alpha\leq1$), then the knowledge distillation Loss can be given by the following formula:-
\begin{equation}
\label{eq:formula2}
\mathcal{L}\textsubscript{k.d.} = \alpha * \mathcal{L}\textsubscript{softmax} + (1 - \alpha) * \mathcal{L}\textsubscript{KL}
\end{equation}

\par The $y_{o, c}$ represents the truth label for that particular sample, and $p_{o, c}$ represents the softmax probabilities obtained after the final fully connected layer (Fc4) $L \in R\textsuperscript{128 x 4}$. Further, the $\hat{y\textsubscript{c}}$ and $y\textsubscript{c}$ are the predicted and the actual probability distributions respectively for a given soccer frame sample.

\end{itemize}

\subsection{Experiments}
\label{sec:exp}
\textbf{Datasets Used.} Here, we use two datasets for our work. The first dataset is the Kinetics 400 dataset which consists of 400 diverse action classes of everyday activities with over 300, 000 videos. We used this dataset for pre-training our backbone model as it can learn more generalized features.
\par For benchmarking our results, we used SoccerDB1. Further details regarding our dataset have already been discussed in detail in Section \ref{sec:db}.

\begin{table}[h]
\centering
\begin{tabular}{|c | c|} 
 \hline
 Alpha ($\alpha$) & Model Accuracy \\ [0.5ex] 
 \hline\hline
 0.95 & 66.31\% \\ 
 \hline
 \textbf{0.90} & \textbf{67.20\%}\\ 
 \hline
 0.97 & 62.8\% \\ [1ex] 
 \hline
\end{tabular}
\caption{Alpha vs Model accuracy comparison}
\label{tbl:alpha}
\end{table}

\textbf{Implementation Details.} We first train the jointnet on the Soccer Dataset for 100 epochs. The batch size was kept to 64 and the Cross Entropy Loss function was used. The Adam optimizer was chosen with a learning rate of 0.0001 and CosineAnnealing rate decay scheduler.
\par Next, we train the student model which is a ResNet-18. We take a batch size of 128 and train the student model for 200 epochs. The KullBack-Liebler divergence loss function was used here alongwith the distillation loss as described in equation 3. Table \ref{tbl:alpha} illustrates the considerable change in accuracy with the changing value of $\alpha$, and we found the optimal value to be 0.90. The value of Temperature ($\tau$) was taken to be 6, SGD Optimizer was used here with a constant learning rate of 0.0001, momentum of 0.9 and a constant weight decay of 5e-4. Then the obtained student model was simply plugged into the evaluation framework to obtain the corresponding accuracy. As evident from Figure \ref{fig:accplot1} in Section \ref{sec:results}, experiments were run for 200 epochs till the validation accuracy started saturating.


\par All the input video frames were resized to $224 \times 224$ RGB images using the spatial cropping strategy as outlined in \cite{chen2021deep}. All the accuracies reported were sampled over 5 runs to ensure the reproducibility of results. All models were trained on a 32 GB NVIDIA V100 GPU.

\section{Experimental Results}
\label{sec:results}

    

\par \textbf{Accuracy Metrics.} All accuracies reported here are Top-1 accuracies. All figures were sampled over 5 runs. The student model is ResNet-18 in all cases. In our setting, we have used the Top-1/Top-5 accuracy metric for evaluation in all our experiments as used in several previous works \cite{girdhar2019distinit}. Top-1 accuracy refers to when the 1\textsuperscript{st} predicted model label matches with the ground truth label for the particular frame. Using this metric, a particular soccer video is considered to be classified correctly only when atleast half or more of it's total frames match with the corresponding ground truth label. Thus, we report the accuracies obtained using various backbones in Table \ref{tbl:top1}.

\begin{table}[]
   
        \centering
        
        \begin{tabular}{|c | c | c|} 
         \hline
         BackBone & Teacher Acc. & Student Acc. \\ [0.5ex] 
         \hline\hline
         ResNet-50 & 60.00\% & 65.26\% \\ 
         \hline
         ResNet-101 & 62.10\% & \textbf{67.20\%} \\ [1ex] 
         \hline
        \end{tabular}
\caption{Top-1 validation accuracies obtained on the soccer dataset using various network backbones.}
\label{tbl:top1}
\end{table}

\par We note that directly using the pre-trained backbone model yields a very poor accuracy of 7.7\% highlighting the need for a generalized network. We also see that the student model, with proper training and sufficient number of epochs exceeds the teacher model in accuracy. 

\par When the fine-tuning dataset is small, it is very difficult to ensure the model does not overfit to the dataset. Here, pre-training is carried out on the Kinetics 400 dataset, which is significantly larger in comparison to our Soccer Dataset. To prevent overfitting, we rigorously apply regularizers such as Batch Normalization on both the TAM and FrontNet module and dropout on the Student Model. However, such concerns may still remain to some extent as highlighted in \cite{carreira2017quo}.

\begin{table}[]
\centering
      \begin{tabular}{|c | c | c|} 
         \hline
         Model & Dataset & Accuracy* \\ [0.5ex] 
         \hline\hline
         Russo et al. \cite{russo2018sports} & 300 & 32.00\% \\ 
         \hline
         Kukleva et al.\textsuperscript{\textdagger} \cite{kukleva2019utilizing} & 4152 & 94.50\% \\ 
         \hline
         SoccerKDNet (R50) & 448 & 65.26\% \\ 
         \hline
         SoccerKDNet (R101) & 448 & \textbf{67.20\%} \\ [1ex] 
         \hline
        \end{tabular}
\caption{Comparison of accuracy of SoccerKDNet with similar methods.}
\label{tbl:cmp}
\end{table}

    

As it can be seen from Table \ref{tbl:cmp}, our model outperforms all other existing models. The SoccerData dataset by Kukleva et al. \cite{kukleva2019utilizing} is an image dataset not video dataset. On that particular dataset, the model requires digit level bounding boxes and human keypoint annotations which our dataset does not have and there are no trained models provided by the authors to be used publicly for testing. Further, our models are trained on video data which cannot be ideally tested on image datasets without compromising on crucial information such as the temporal sequence present in a video.

\par Several earlier works such as Two Stream Networks \cite{feichtenhofer2016convolutional} have millions of parameters and hence are unsuitable for edge deployment. Using knowledge distillation here, we show that even simple 2D networks such as ResNet-18 can be used for action classification. Table \ref{tbl:alpha} highlights this aspect by listing the networks used in our work - all of them have fewer than 50 mil. parameters and the backbone network is frozen. For context, 3D networks such as C3D\cite{tran2015learning} has 73 million parameters.

\begin{figure}[h]
		\centering
            \qquad
            \includegraphics[height=0.3\linewidth, width=0.9\linewidth,trim=4 4 4 1,clip]{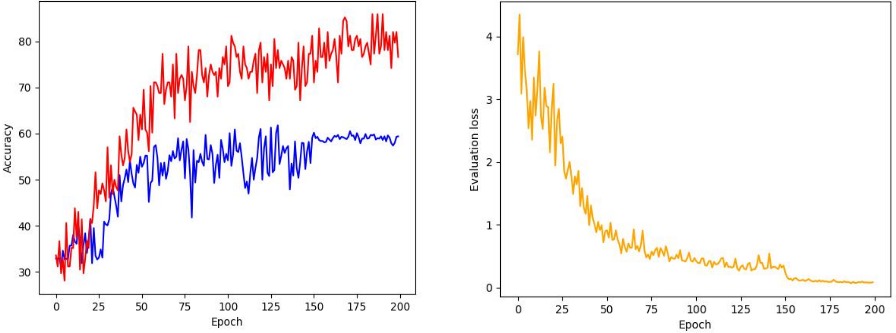}
         
		\caption{On the left: red and blue curves denote the train and validation accuracies respectively. On the right, the validation loss is shown.}
\label{fig:accplot1} 
	\end{figure}
	
From Figure \ref{fig:accplot1}, it can be seen that the student Model is able to achieve a Train accuracy as high as 77.9\% and validation accuracies above 60\% underscoring the generalizability and effectiveness of our proposed network. On the right, the corresponding validation loss curve obtained using a ResNet-18 student and a ResNet-50 teacher model is shown.



\subsection{Ablation Study}
We also perform a mini ablation study on a variety of factors: the backbone network, stage in which distillation is applied, layers in frontnet module and various hyperparameters.
\begin{itemize}
\item \textit{Backbone Network:} As it can be seen from Table \ref{tbl:top1}, we experimented with 2 backbone networks. The TAM-ResNet101 backbone model performs the best for both the teacher and student models. Further, employing parameter heavy models such as ResNet-152 as the backbone would not only defeat the purpose of moving towards a more online solution but the relatively small size of the target dataset for fine-tuning means there will be considerable concerns in performance due to model over-fitting.
\item \textit{Distillation Stage:} There are two possibilities: directly distill the frozen TAM backbone module and then use the distilled model as a plug-in within the network. We call this as the early distillation where the distillation is done early on in the network. However, performance using this approach is not satisfactory: taking the TAM-ResNet50 as the backbone Teacher model, we get only 8.3\% accuracy on the Kinetics400 dataset using the ResNet-18 as the student model. We suspect this is due to the inability of the student model to directly learn features from the heavy teacher model. Therefore, we chose to go with the late stage distillation process.
\item \textit{FrontNet Module and Hyperparameters:} We found adding dropout layers decreased the accuracy by 1.2\%, adding more fully connected layers in the latter half of the FrontNet module decreased accuracy as much as 4.9\% of the teacher module to 57.2\%. We did not find much difference in accuracy on using NLL loss instead of CrossEntropy so we chose to keep it. We found a learning rate of 0.0001 and batch size of 64 and 128 to be optimal for all our experiments as bigger batches lead to better performance.
\end{itemize}

\section{Conclusion}
\par In this paper, we introduce a new Soccer Dataset consisting of 4 diverse classes and over 70 video clips per class. The network not only provides the flexibility of using any of the original (teacher) model as the classifier but also has the option of using a smaller network (ResNet-18 in our case) as the student network. In future, we plan to add more action classes in our dataset. Also we plan to utilize SoccerKDNet for soccer event detection based on the actions of the players.

\bibliographystyle{splncs04}
\bibliography{ref}
\end{document}